\documentclass[conference]{IEEEtran}
\IEEEoverridecommandlockouts
\usepackage{cite}
\usepackage{amsmath,amssymb,amsfonts}
\usepackage{algorithmic}
\usepackage{graphicx}
\usepackage{textcomp}
\usepackage{xcolor}
\usepackage{tabularx}
\usepackage{booktabs}
\usepackage{longtable}
\usepackage{url}
\usepackage{array}
\usepackage{longtable}
\usepackage{lscape}
\usepackage{diagbox}
\usepackage{multirow}

\def\BibTeX{{\rm B\kern-.05em{\sc i\kern-.025em b}\kern-.08em
    T\kern-.1667em\lower.7ex\hbox{E}\kern-.125emX}}
\begin{document}
\title{A survey on Graph Deep Representation Learning for Facial Expression Recognition}
\author{\IEEEauthorblockN{Théo Gueuret}
\IEEEauthorblockA{\textit{Univ. Lille, CNRS, Centrale Lille} \\
\textit{UMR 9189 CRIStAL}\\
Lille, France \\
theo.gueuret@univ-lille.fr}
\and
\IEEEauthorblockN{Akrem Sellami}
\IEEEauthorblockA{\textit{Univ. Lille, CNRS, Centrale Lille} \\
\textit{UMR 9189 CRIStAL}\\
Lille, France \\
akrem.sellami@univ-lille.fr}
\and
\IEEEauthorblockN{Chaabane Djeraba}
\IEEEauthorblockA{\textit{Univ. Lille, CNRS, Centrale Lille} \\
\textit{UMR 9189 CRIStAL}\\
Lille, France \\
chabane.djeraba@univ-lille.fr}
}
\newcolumntype{R}[1]{>{\raggedleft\arraybackslash }b{#1}}
\newcolumntype{L}[1]{>{\raggedright\arraybackslash }b{#1}}
\newcolumntype{C}[1]{>{\centering\arraybackslash }b{#1}}
\maketitle

\begin{abstract}
This comprehensive review delves deeply into the various methodologies applied to facial expression recognition (FER) through the lens of graph representation learning (GRL). Initially, we introduce the task of FER and the concepts of graph representation and GRL. Afterward, we discuss some of the most prevalent and valuable databases for this task. We explore promising approaches for graph representation in FER, including graph diffusion, spatio-temporal graphs, and multi-stream architectures. Finally, we identify future research opportunities and provide concluding remarks.
\end{abstract}

\section{Introduction}
Facial expressions transcend cultural barriers, playing a crucial role in communication and human interaction. They serve as a natural conduit for expressing emotions and sharing information, making facial expression recognition (FER) essential for deciphering emotional subtleties and anticipating reactions across various contexts. FER finds application in a multitude of sectors including healthcare, education, the automotive industry, marketing, robotics, entertainment, and customer service, underscoring its significance and potential for innovation. Historically, the development of FER technologies has been dominated by deep learning (DL) techniques, notably convolutional neural networks (CNN).

Despite notable progress, these methods face significant challenges in accurately modeling the complexity of facial expressions. These challenges include locating faces in cluttered environments, variations in lighting that can obscure or distort essential facial features, and analysis of face texture, with repetitive patterns or skin peculiarities. We can also mention occlusion problems, where external elements hide parts of the face, expression differences among individuals, and varied head poses. Furthermore, efficiently encoding faces, poses a crucial challenge, necessitating robust representations. 
Finally, the debate between discrete and continuous emotion representation, as researchers have to choose between the precision of discrete categories and the nuance of continuous models to classify the human emotions. These challenges represent significant barriers to the generalization of FER models, highlighting the importance of approaches capable of handling the intrinsic variability of human expressions and image capture conditions.

These challenges have opened the search for new approaches, among which graph representation learning (GRL) stands out as a promising solution. Leveraging relational and structural data, this method offers an innovative technique to overcome the challenges of FER and opens a new direction for advancements in the field. Overcoming all obstacles will require continuous innovations in image processing, data modeling, and machine learning, emphasizing the importance of future research to explore new methods aimed at improving the accuracy, robustness, and efficiency of FER systems.
\subsection{Facial Expression Recognition (FER)}
FER is both a complex and comprehensive task, typically associated with a classification problem. Its history is also rooted in the disciplines of psychology and neuroscience. Its origins can be traced back to the pioneering work "The Expression of the Emotions in Man and Animals" by C. Darwin \cite{darwin_1872}, in which he argued for the universality of some facial expressions among human beings.

\begin{figure}[htbp]
\centerline{\includegraphics[width=0.90\columnwidth]{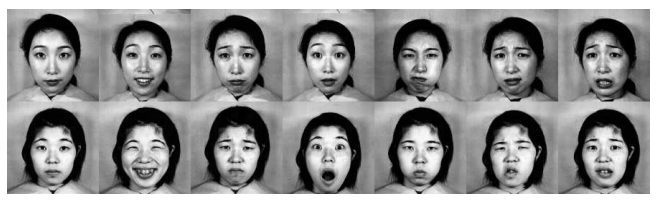}}
\caption{Samples from the JAFFE \cite{jaffe_1998} FER database.}
\label{fig:jaffe}
\end{figure}

Throughout the 20th century, the field had a large growth due to the introduction of the Facial Action Coding System (FACS) by P. Ekman \textit{et al.} \cite{facs_1978}. This system, which focused on classifying facial expressions through action units (AUs), marked a major turning point, significantly enriching research and applications in this area. At the end of the 20th century, its integration into computing allowed the automatization of processes. This integration allowed major innovations in machine learning \cite{cootes1995active}, such as the tracking of facial landmarks. New databases specialized in emotions appeared at the end of the century, we can cite the Japanese Female Facial Expression (JAFFE) database by M. Lyons \textit{et al.} \cite{jaffe_1998}, presented in Fig. \ref{fig:jaffe}, and the Cohn-Kanade dataset by T. Kanade \textit{et al.} \cite{ck_2000}. Their release made possible the development of data-driven approaches as we know them today.

More recently, DL marked an important stepping stone for the task of FER \cite{fathallah2017facial}. The large adoption of CNNs not only improved the accuracy and robustness of FER systems but also extended their fields of application, facilitating their integration into diverse areas of applications such as mental health diagnostics, or human-machine interaction. The research on the FER task has become increasingly dynamic and the results follow a rapid progression with more challenging databases to align with the recent technological advancements.
\subsection{Graph Theory}
GRL originally comes from graph theory, a mathematical discipline initiated by L. Euler \cite{euler_1741} with his solution to the Königsberg bridge problem.
\begin{figure}[htbp]
\centerline{\includegraphics[width=0.67\columnwidth]{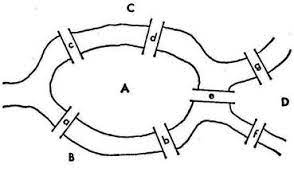}}
\caption{Königsberg bridge problem \cite{pasqualini2024}.}
\label{fig:ponts}
\end{figure}

This breakthrough laid the foundations for graph theory. The beginnings of neural networks encouraged researchers to use graph theory for their experiments, but the computational capabilities of the early models rapidly limited the models to structured data, setting aside graph-structured data.
New techniques then allowed more application of graph theory to machine learning. Techniques such as spectral clustering \cite{von2007tutorial}, utilized the eigenvalues of graph laplacians, offering modern tools for deeper analysis and processing of graph-structured data.

The early 2000s saw the introduction of graph neural networks (GNNs) \cite{scarselli_2008}, thus greatly improving GRL, the combination of neural networks to graph data opened new paths of research for graph machine learning. Their development led quickly to graph convolutional networks (GCNs) \cite{kipf2016semi}, and graph attention mechanisms \cite{velickovic_2017}, greatly improving GCNs effectiveness. GRL became dynamic through the years, rapidly expanding due to its versatile applications. New methods are explored to improve the models' efficiency and scalability.

This article aims to provide a general overview of GRL approaches for FER. Following this introduction, we will discuss graphs representation in section 2, the GRL in section 3, the main databases in section 4, recent and promising approaches in section 5, the research opportunities in section 6, and finally the conclusion in section 7.
\section{Graph representation}
Graphs play a crucial role in modeling complex relational structures. A graph can be defined in the form $G = (X, E)$
where $X$ represents a finite, non-empty set of elements called vertices, and $E$ denotes a set of unordered pairs of elements from $X$ named edges, such that $E$ is a subset of $\{\{x_{i},x_{j}\} \in X^2 | x_{i} \neq x_{j}\}$
In the context of directed graphs, an edge, then called an arc, is represented by an ordered pair such that $E$ is a subset of $\{(x_{i},x_{j}) \in X^2 | x_{i} \neq x_{j}\}$
indicating the direction of the relationship from $x_{i}$ to $x_{j}$.

A variant of this structure is the weighted graph, considered for both directed and undirected graphs. It is described by the form $G = (X, E, \omega)$ where each edge, or arc $\{x_{i}, x_{j}\} \in E$ is associated with a real number $\omega(e) | e \in E$ by a function $\omega: E \rightarrow \mathbf{R}$, thus forming a triplet $(X, E, \omega)$. This weighting allows the addition of additional information such as cost, distance, or another relevant metric, making weighted graphs particularly useful for representing complex structures in various contexts such as traffic optimization or social network analysis. In the case of weighting by similarity, the weight $\omega_{i,j}$ for the edge $e_{i,j}$ is defined as follows:
\[
\omega_{i,j} = \text{dist}(x_{i}, x_{j}) = \lVert x_{i} - x_{j} \rVert_{2}
\]
where $x_{i}$ and $x_{j}$ are the characteristics of the vertices $x_{i}$ and $x_{j}$, which can represent geometric coordinates, feature vectors from machine learning or filters (e.g., Gabor, EMPs).
The adjacency matrix $\mathbf{A}$ can thus be formed as follows:
\[
A_{i,j}= \left\{
\begin{array}{ll}
1 & \text{if } w_{i,j} < s, \\
0 & \text{otherwise}.
\end{array}
\right.
\]
or:
\[
A_{i,j}= \left\{
\begin{array}{ll}
w_{i,j} & \text{if } w_{i,j} < s, \\
0 & \text{otherwise}.
\end{array}
\right.
\]
The threshold $s$, arbitrarily defined, plays a crucial role in determining the graph's structure by influencing the connectivity between vertices.

The creation of the adjacency matrix and the formation of weights leave much room for creativity, with numerous possible and imaginable methods reflecting the robustness and generalization capabilities of graphs.

\begin{figure}[htbp]
\centerline{\includegraphics[scale=0.23]{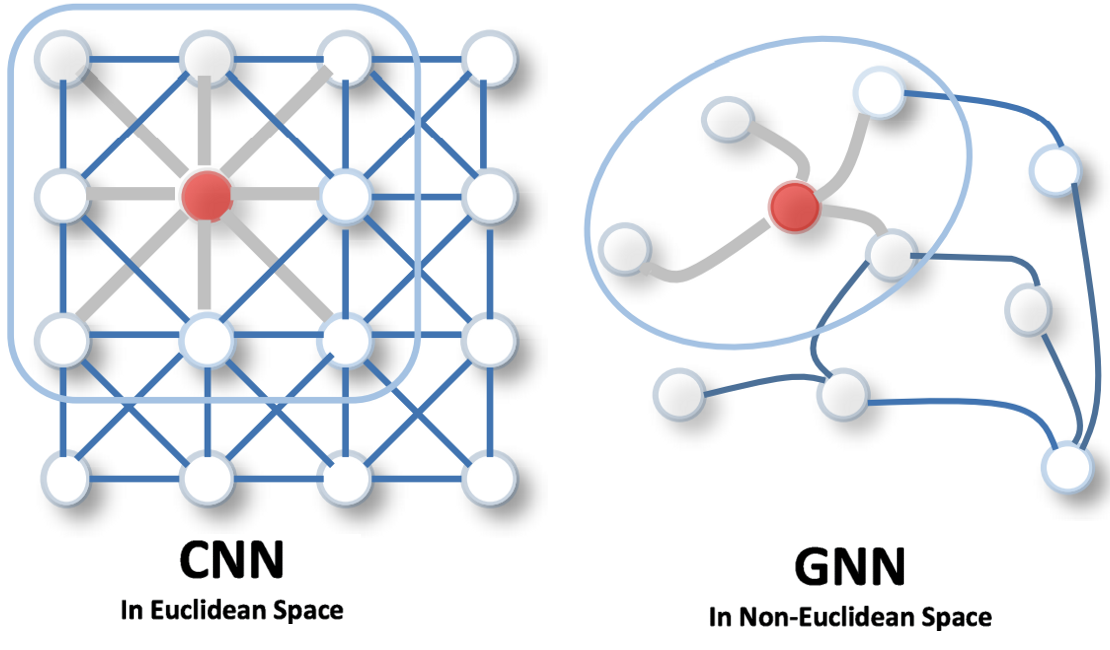}}
\caption{Difference between CNN and GNN \cite{wu2020}}
\label{fig:comparaison}
\end{figure}

As shown in Fig. \ref{fig:comparaison}, usual DL methods use region grids. In contrast, graph-based DL uses nodes and edges. A node can represent a facial landmark, a region, or a sample. An edge can express the relationship between two nodes with the same annotation, that is, the same expression (anger, sadness, etc.), or the same semantic region like the eye, nose, mouth, etc. It can also express the spatial relationship between regions (eyes, nose, mouth). Finally, it can represent the temporal relationship of the same region at different times.

In this article, we will discuss several types of graphs representing various structures. The nodes of these graphs can be of different natures, such as facial landmarks, pixels, action units, or regions of interest. In some cases, nodes may represent entire samples of images and videos. As for the edges, these can be weighted according to the Euclidean distance of facial landmarks, the similarity of node features, temporality (between images of a video), or just represent the structure of the face via links between facial landmarks. This variety demonstrates the breadth and flexibility of graphs in solving the task of FER.
\section{Graph Representation Learning (GRL)}
GRL aims to map the elements of a graph, or subgraphs into continuous vector spaces while preserving the graph's inherent structural and attribute information. Formally, consider a graph \( G = (X, E) \) where \( X \) is the set of vertices and \( E \) is the set of edges. The goal of GRL is to learn a function \( f: X \rightarrow \mathbb{R}^d \), where \( d \) is the dimension of the embedding space and \( f(x) \) represents the vector embedding of the node \( x \in X \). This function \( f \) should capture both local properties (e.g., node features, adjacency) and global properties (e.g., community structure, connectivity patterns) of the graph. Mathematically, the optimization objective often involves minimizing a loss function \( \mathcal{L} \) that measures the discrepancy between the actual graph structure and the structure predicted from the embeddings. For instance, in node classification, the loss function can be defined as the cross-entropy between the predicted node labels and the true labels. In random walk-based methods, the loss function might be the negative log-likelihood of observing a node given its context nodes in the random walks. In GNNs, the embedding of a node \( v \) is iteratively updated by aggregating the embeddings of its neighbors, typically using a neural network parameterized by weights \( \theta \), which are learned during the training process. Mathematically, this can be expressed as: 
\[
h_v^{(k)} = \sigma \left( \sum_{u \in \mathcal{N}(v)} \text{AGG}(h_u^{(k-1)}, h_v^{(k-1)}; \theta) \right),
\]
where \( h_v^{(k)} \) is the embedding of node \( v \) at the \( k \)-th layer, \( \mathcal{N}(v) \) denotes the set of neighbors of \( v \), \(\text{AGG} \) is an aggregation function (e.g., mean, sum, attention), and \( \sigma \) is a non-linear activation function. The overall learning process involves jointly optimizing the parameters \( \theta \) to ensure that the learned embeddings effectively capture the desired properties of the graph.
\section{Databases}
The databases for the FER task can be categorized into two distinct groups: static and dynamic. These databases also differ based on the context of data collection: they can include 'in-lab' subjects, characterized by controlled experimental conditions, or 'in-the-wild' samples, used for their complex and uncontrolled environments, often leading to a decrease in the accuracy of predictive models. Furthermore, some datasets adopt a hybrid approach, merging 'in-lab' and 'in-the-wild' data to create a database that combines the simplicity of controlled environments with the complexity of natural settings. We will briefly present the following databases: JAFFE \cite{jaffe_1998}, CK+ \cite{ck_2000}, CASME II \cite{casme2014}, SAMM \cite{samm2016}, SMIC \cite{smic2013}, SFEW 2.0 \cite{sfew}, BU-3DFE \cite{bu-3dfe}, KDEF \cite{kdef}, RAF-DB \cite{raf-db}, FER-2013 \cite{fer2013}, ExpW \cite{expw}, DISFA \cite{disfa}, AffectNet \cite{affectnet}, EmotioNet \cite{emotionet}, AM-FED \cite{am-fed}, Oulu-CASIA \cite{casia}, BU-4DFE \cite{bu-4dfe}, SASE-FE \cite{sase-fe}, EmoReact \cite{emoreact}, SNAP-2DFE \cite{snap-2dfe}, AFEW \cite{afew}, MUG \cite{mug}, MMI \cite{mmi}, DFEW \cite{dfew}, MAFW \cite{mafw2022}, and FERV39K \cite{ferv39k} in Table \ref{table:fer_datasets}. In this table, "AUs" stands for Action Units. These are specific facial muscle movements that are coded in the Facial Action Coding System (FACS). When a dataset lists "AUs" under the "classes" column, it indicates that the dataset is annotated with these detailed facial muscle movements instead of or in addition to more general emotional categories. 

Generally speaking, deep graph learning approaches are being tested on static and dynamic databases of macro-expressions. This is the case for all presented datasets, with the three exceptions of SAMM, SMIC, and CASME II, which include micro-expressions.
\begin{table}[!ht]
\centering
\caption{Summary of different static and dynamic datasets for the fer task}
\scriptsize 
\begin{tabularx}{\columnwidth}{|L{1.5cm}|C{1cm}|C{1.1cm}|C{1.2cm}|C{0.8cm}|C{0.7cm}|}
\hline
\textbf{Dataset} & \textbf{\# samples} & \textbf{Individuals} & \textbf{Expression type} & \textbf{Context} & \textbf{\# classes} \\
\hline
\multicolumn{6}{|c|}{Static datasets} \\
\hline
JAFFE & 213 & 10 & macro & Lab & 7 \\
CK+ & 593 & 123 & macro & Lab & 7 \\
SFEW 2.0 & 1,766 & N/A & macro & Wild & 7 \\
BU-3DFE & 2,500 & 100 & macro & Lab & 6 \\
KDEF & 4,900 & 70 & macro & Lab & 7 \\
RAF-DB & 29,672 & N/A & macro & Wild & 11 \\
FER-2013 & 35,887 & N/A & macro & Wild & 7 \\
ExpW & 91,793 & N/A & macro & Wild & 7 \\
DISFA & 130,815 & 27 & macro & Lab & AUs \\
AffectNet & 1,000,000 & N/A & macro & Wild & 8 \\
EmotioNet & 1,000,000 & N/A & macro & Wild & AUs \\
\hline
\multicolumn{6}{|c|}{Dynamic datasets} \\
\hline
SAMM & 159 & 32 & micro/macro & Lab & AUs \\
SMIC & 164 & 16 & micro & Lab & AUs \\
AM-FED & 242 & N/A & macro & Wild & 17 \\
CASME II & 247 & N/A & micro & Lab & AUs \\
CK+ & 593 & 123 & macro & Lab & 7 \\
BU-4DFE & 606 & 101 & macro & Lab & 6 \\
SASE-FE & 643 & 54 & macro & Lab & 6 \\
EMOREACT & 1102 & N/A & macro & Wild & 17 \\
SNAP-2DFE & 1260 & 15 & macro & Lab & 7 \\
AFEW & 1,426 & 330 & macro & Wild & 7 \\
MUG & 1462 & 77 & macro & Lab & 6 \\
Oulu-CASIA & 2880 & 80 & macro & Lab & 6 \\
MMI & 2,900 & 75 & macro & Lab & AUs \\
MAFW & 10,045 & N/A & macro & Wild & 11 \\
DFEW & 16,372 & N/A & macro & Wild & 7 \\
FERV39K & 39,546 & N/A & macro & Wild & 7 \\
\hline
\end{tabularx}
\label{table:fer_datasets}
\end{table}

\section{Recent Methodologies}
\begin{table*}[!ht]
\caption{Comparative study}
\begin{center}
\begin{tabular}{|L{2.8cm}|C{1.4cm}|C{5.5cm}|C{5.0cm}|}
\hline
Reference & Method & Features & Challenges \& Limits \\
\hline
J. Zhou \textit{et al.} \cite{zhou2020learning} 
& SGCN
& \begin{tabular}[c]{@{}c@{}c@{}}\textbf{nodes}: facial landmarks\\\textbf{edges}: situational links\\\textbf{features}: HOG \& coordinates\end{tabular}
& \begin{tabular}[c]{@{}c@{}c@{}c@{}}Robustness in diverse conditions\\Dynamic learning of the edges\\Generalisation to other datasets\\Computational complexity\end{tabular}
\\
\hline
J. Zhou \textit{et al.} \cite{zhou2020facial} 
& STSGN 
& \begin{tabular}[c]{@{}c@{}c@{}}\textbf{nodes}: facial landmarks\\\textbf{edges}: structural \& temporal\\
\textbf{features}: HOG and coordinates\end{tabular}
& \begin{tabular}[c]{@{}c@{}c@{}c@{}}Dynamical facial topology\\Interpretability of the generalisation\\Topology optimisation\\GCN improvement\end{tabular}
\\
\hline
T. Chen \textit{et al.} \cite{chen2021cross} 
& Cross-Domain
& \begin{tabular}[c]{@{}c@{}c@{}}\textbf{nodes}: region of interest\\\textbf{edges}: intra/inter domain\\\textbf{features}: holistic and local\end{tabular}
& \begin{tabular}[c]{@{}c@{}c@{}c@{}c@{}}Intra-classes variability\\Robust features learnings\\Real-Time processing\\Dependancy to learning data\\Difficulties linked to spatio-temporality\end{tabular}
\\
\hline
L. Lei \textit{et al.} \cite{lei2021micro}
& Dual-Stream
& \begin{tabular}[c]{@{}c@{}c@{}}\textbf{nodes}: patches of facial landmarks\\\textbf{edges}: learned via attention\\\textbf{features}: deep via convolution\end{tabular}
& \begin{tabular}[c]{@{}c@{}c@{}}Micro/Macro-expressions\\AUs Integration\\Generalisation to other datasets\end{tabular}
\\
\hline
A. Panagiotis \textit{et al.} \cite{panagiotis2021exploiting} 
& Dual-Stream 
& \begin{tabular}[c]{@{}c@{}c@{}}\textbf{nodes}: 7 emotions + VA\\\textbf{edges}: by correlation\\\textbf{features}: deep via DenseNet\end{tabular}
& \begin{tabular}[c]{@{}c@{}c@{}}"In-the-Wild" data\\MTL model\\Capturing features\end{tabular}
\\
\hline
X. Jin \textit{et al.} \cite{jin2021learning} 
& DDRGCN
& \begin{tabular}[c]{@{}c@{}c@{}}\textbf{nodes}: region of interest\\\textbf{edges}: structural\\\textbf{features}: deep by auto-encoder\end{tabular}
& \begin{tabular}[c]{@{}c@{}c@{}}Pre-definition of the adjacency matrix\\Real-Time application\\Modelling spatial dependencies\end{tabular}
\\
\hline
R. Zhao \textit{et al.} \cite{zhao2022spatial}
& Dual-Stream
& \begin{tabular}[c]{@{}c@{}c@{}}\textbf{nodes}: patches of facial landmarks\\\textbf{edges}: structural\\\textbf{features}: geometrical via transformers\end{tabular}
& \begin{tabular}[c]{@{}c@{}c@{}c@{}}Use of spatio-temporal data\\Use of Attention and Transformers\\Use of "In-lab" data only\end{tabular}
\\
\hline
N. Xie \textit{et al.} \cite{xie2023} 
& GSTG
& \begin{tabular}[c]{@{}c@{}c@{}}\textbf{nodes}: facial landmarks\\\textbf{edges}: structural\\\textbf{features}: deep by convolution\end{tabular}
& \begin{tabular}[c]{@{}c@{}c@{}}\\Use of Attention and Transformers\\Dynamic expression modeling\\Generalisation to other datasets\end{tabular}
\\
\hline
R. Wang \textit{et al.} \cite{wang2024graph}
& GDRL
& \begin{tabular}[c]{@{}c@{}c@{}}\textbf{nodes}: dataset's samples\\\textbf{edges}: relation between samples\\\textbf{features}: deep by réduction\end{tabular}
& \begin{tabular}[c]{@{}c@{}c@{}c@{}}Data variations\\Graph diffusion mechanism\\Learning invariant representations\\Generalisation to other domains\\Dependancy to adaptation technique\end{tabular}
\\
\hline
F. Wang \textit{et al.} \cite{wang2024dynamic} 
& DSGCN
& \begin{tabular}[c]{@{}c@{}c@{}}\textbf{nodes}: videos' frames \& \\ dataset's samples\\\textbf{edges}: frames similarity \& \\ embedding similarity \\\textbf{features}: geometrical by spatial transformer \& \\ spatio-temporal by temporal transformer\end{tabular}
& \begin{tabular}[c]{@{}c@{}c@{}c@{}}Capture of spatio-temporal relationships\\Robustness to "In-the-Wild" data\\Using Attention and Transformers\\Adaptation to greater data variations\end{tabular}
\\
\hline
\end{tabular}
\label{table:etude_shortened}
\end{center}
\end{table*}

\begin{table*}[!ht]
\caption{Comparative study of graph-based and classical methods for dynamic facial expression recognition across various datasets}
\begin{center}
\scriptsize
\renewcommand{\arraystretch}{1.4} 
\setlength{\tabcolsep}{4pt} 
\begin{tabular}{|c|c|c|c|c|c|c|c|c|c|c|c|c|c|c|c|c|c|c|c|c|c|}
\hline
\rotatebox[origin=c]{90}{\textbf{Type}} & \diagbox[width=1.5cm,height=2.5cm]{\textbf{Methods}}{\rotatebox[origin=rB]{90}{\textbf{Datasets}}} 
 & \rotatebox[origin=rB]{90}{\textbf{CK+}} & \rotatebox[origin=cB]{90}{\textbf{Oulu-CASIA}} & \rotatebox[origin=cB]{90}{\textbf{AFEW}} & \rotatebox[origin=cB]{90}{\textbf{JAFFE}} & \rotatebox[origin=cB]{90}{\textbf{SFEW 2.0}} & \rotatebox[origin=cB]{90}{\textbf{FER2013}} & \rotatebox[origin=cB]{90}{\textbf{ExpW}} & \rotatebox[origin=cB]{90}{\textbf{CASME II}} & \rotatebox[origin=cB]{90}{\textbf{SAMM}} & \rotatebox[origin=cB]{90}{\textbf{AffectNet}} & \rotatebox[origin=cB]{90}{\textbf{Aff-Wild2}} & \rotatebox[origin=cB]{90}{\textbf{RaFD}} & \rotatebox[origin=cB]{90}{\textbf{RAF-DB}} & \rotatebox[origin=cB]{90}{\textbf{eNTERFACE05}} & \rotatebox[origin=cB]{90}{\textbf{CAER}} & \rotatebox[origin=cB]{90}{\textbf{KDEF}} & \rotatebox[origin=cB]{90}{\textbf{TFEID}} & \rotatebox[origin=cB]{90}{\textbf{FERV39K}} &
 \rotatebox[origin=cB]{90}{\textbf{DFEW}} & \rotatebox[origin=cB]{90}{\textbf{MAFW}} \\
\hline
\multirow{10}{*}{{\rotatebox[origin=rB]{90}{\textbf{Graph-based}}}}
& \textbf{SGCN} \cite{zhou2020learning} & 98.9 & 87.5 & 45.1 & $\times$ & $\times$ & $\times$ & $\times$ & $\times$ & $\times$ & $\times$ & $\times$ & $\times$ & $\times$ & $\times$ & $\times$ & $\times$ & $\times$ & $\times$ & $\times$ & $\times$ \\
& \textbf{STSGN} \cite{zhou2020facial} & 98.6 & 87.2 & $\times$ & $\times$ & $\times$ & $\times$ & $\times$ & $\times$ & $\times$ & $\times$ & $\times$ & $\times$ & $\times$ & $\times$ & $\times$ & $\times$ & $\times$ & $\times$ & $\times$ & $\times$ \\
& \textbf{Cross-Domain} \cite{chen2021cross} & 85.3 & $\times$ & $\times$ & 61.5 & 56.4 & 58.9 & 68.5 & $\times$ & $\times$ & $\times$ & $\times$ & $\times$ & $\times$ & $\times$ & $\times$ & $\times$ & $\times$ & $\times$ & $\times$ & $\times$ \\
& \textbf{Dual-Stream} \cite{lei2021micro} & $\times$ & $\times$ & $\times$ & $\times$ & $\times$ & $\times$ & $\times$ & 74.3 & 74.3 & $\times$ & $\times$ & $\times$ & $\times$ & $\times$ & $\times$ & $\times$ & $\times$ & $\times$ & $\times$ & $\times$ \\
& \textbf{Dual-Stream} \cite{panagiotis2021exploiting} & $\times$ & $\times$ & $\times$ & $\times$ & $\times$ & $\times$ & $\times$ & $\times$ & $\times$ & 66.5 & 48.9 & $\times$ & $\times$ & $\times$ & $\times$ & $\times$ & $\times$ & $\times$ & $\times$ & $\times$ \\
& \textbf{DDRGCN} \cite{jin2021learning} & 94.3 & 73.3 & $\times$ & $\times$ & $\times$ & $\times$ & $\times$ & $\times$ & $\times$ & $\times$ & $\times$ & 94.5 & 58.3 & $\times$ & $\times$ & $\times$ & $\times$ & $\times$ & $\times$ & $\times$ \\
& \textbf{Dual-Stream} \cite{zhao2022spatial} & 98.8 & 89.2 & 51.2 & $\times$ & $\times$ & $\times$ & $\times$ & $\times$ & $\times$ & $\times$ & $\times$ & $\times$ & $\times$ & 54.6 & 77.0 & $\times$ & $\times$ & $\times$ & $\times$ & $\times$ \\
& \textbf{GSTG} \cite{xie2023} & $\times$ & 88.4 & 55.8 & $\times$ & $\times$ & $\times$ & $\times$ & $\times$ & $\times$ & $\times$ & $\times$ & $\times$ & $\times$ & $\times$ & $\times$ & $\times$ & $\times$ & $\times$ & $\times$ & $\times$ \\
& \textbf{GDRL} \cite{wang2024graph} & 71.9 & $\times$ & $\times$ & 61.9 & $\times$ & 46.9 & $\times$ & $\times$ & $\times$ & $\times$ & $\times$ & $\times$ & 55.6 & $\times$ & $\times$ & 78.4 & 62.9 & $\times$ & $\times$ & $\times$ \\
& \textbf{DSGCN} \cite{wang2024dynamic} & $\times$ & $\times$ & 70.6 & $\times$ & $\times$ & $\times$ & $\times$ & $\times$ & $\times$ & $\times$ & $\times$ & $\times$ & $\times$ & $\times$ & $\times$ & $\times$ & $\times$ & 59.9 & 65.5 & $\times$ \\
\hline
\multirow{3}{*}{{\rotatebox[origin=rB]{90}{\textbf{Classical}}}}
& \textbf{MMA-DFER} \cite{chumachenko2024mma} & $\times$ & $\times$ & $\times$ & $\times$ & $\times$ & $\times$ & $\times$ & $\times$ & $\times$ & $\times$ & $\times$ & $\times$ & $\times$ & $\times$ & $\times$ & $\times$ & $\times$ & $\times$ & 77.5 & 58.5 \\
& \textbf{A3lign-DFER} \cite{tao20243} & $\times$ & $\times$ & $\times$ & $\times$ & $\times$ & $\times$ & $\times$ & $\times$ & $\times$ & $\times$ & $\times$ & $\times$ & $\times$ & $\times$ & $\times$ & $\times$ & $\times$ & 51.8 & 74.2 & 53.2 \\
& \textbf{SlowR50-SA} \cite{neshov2024slowr50} & $\times$ & $\times$ & $\times$ & $\times$ & $\times$ & $\times$ & $\times$ & $\times$ & $\times$ & $\times$ & $\times$ & $\times$ & $\times$ & $\times$ & $\times$ & $\times$ & $\times$ & 49.3 & 69.9 & $\times$ \\
\hline
\end{tabular}
\label{table:accuracy_comp}
\end{center}
\end{table*}
There are numerous ways to handle the FER task with graphs, this section will introduce and describe some of the most recent and promising approaches. 

\subsection{Graph diffusion}
Graph diffusion is a technique used in GRL to propagate information across nodes in a graph, the objective is to create edges between nodes beyond direct pairwise similarities. In FER, graph diffusion is usually employed to improve the representation of facial expressions across different domains. Most articles using this approach use graph diffusion to achieve more robust, domain-invariant models. They balance both local and global features and their interconnections. Concretely, this approach usually consists of creating at least two graphs, one for the source dataset (labeled) and one for the target dataset (unlabeled). It then uses the correlation of different types of features, for example, holistic and local, to create a single robust graph that is able to adapt to different domains. 

With "Cross-Domain Facial Expression Recognition: A Unified Evaluation Benchmark and Adversarial GRL," T. Chen \textit{et al.} \cite{chen2021cross}, address the issue of data divergence across different FER datasets. They propose an architecture based on adversarial learning between source/target datasets combined with the creation of intra/inter-domain graphs. The intra-domain and inter-domain graphs are constructed with nodes representing facial regions and connections indicating relationships within and between domains. These graphs are then passed through GCNs that diffuse features to adapt and integrate holistic and local features, addressing domain shifts, thus achieving more robust models. 

With the same intent, R. Wang \textit{et al.} \cite{wang2024graph}, with "Graph-Diffusion-Based Domain-Invariant Representation Learning for Cross-Domain Facial Expression Recognition," also aims to address the issue of data divergence across different FER datasets. The proposed model, based on graph diffusion for domain-invariant representation learning, uses a combination of low-dimensional space representation, local graph embedding, and affinity graph diffusion. The initial affinity graph is constructed using a Gaussian kernel to measure sample similarities. This affinity graph undergoes a diffusion process where similarities are propagated to update the graph, effectively capturing complex, high-order relationships among samples. The diffusion process results in an adaptive affinity graph that reflects not only similar samples but also deeper structural relationships within the data, thus improving recognition performance for FER tasks.

\subsection{Spatio-temporal graphs}
In the context of dynamic data analysis, spatio-temporal aspects represent both a modality and a challenge to be considered. The graphs implementing this aspect should capture both the spatial and temporal features. These graphs often consist of nodes and edges, where nodes represent spatial entities (such as facial landmarks in video-based FER) and edges represent spatial or temporal connections between these entities over time. By incorporating both spatial and temporal dimensions, these graphs enable the modeling of complex interactions and changes within the data, providing a rich data representation for the task. 

J. Zhou \textit{et al.} \cite{zhou2020facial}, with "Facial Expression Recognition Using Spatial-Temporal Semantic Graph Network" introduce the Spatial-Temporal Semantic Graph Network (STSGN), an architecture that leverages the facial topological structure to learn spatial and temporal features, achieving competitive results on popular databases. This approach involves preprocessing facial videos to extract facial landmarks and then constructing a graph representation that captures both spatial structures and temporal dynamics of facial expressions. A GCN is then used to learn and recognize expressions from this complex representation. 

Similarly, and more recently, F. Wang \textit{et al.} \cite{wang2024dynamic}, with "Dynamic-Static Graph Convolutional Network for Video-Based Facial Expression Recognition" propose an architecture named Dynamic-Static Graph Convolutional Network (DSGCN). This model consists of two main elements: the Static Relational Graph (SRG) aimed at learning geometric relationships within a video, and the Dynamic Relational Graph (DRG), which is responsible for creating relationships between different video-graphs of the same batch. This approach enables robust expression recognition while accounting for spatio-temporal information and exploiting similarities between different video samples in a batch, thus improving the diversity of features and the robustness of the model. This architecture achieves state-of-the-art results on several datasets.
\subsection{Dual-Stream graphs}
Another pertinent approach for GRL is the dual-stream architecture.
These architectures typically involve two parallel processing streams that independently learn different aspects or modalities of the data before merging their outputs for a final prediction. The dual-stream approach leverages the strengths of each stream, often leading to improved performance in FER task, where both spatial relationships (captured by graph representations) and specific features (captured by additional modalities) are crucial. 

L. Lei \textit{et al.} \cite{lei2021micro}, with their contribution "Micro-expression Recognition Based on Facial GRL and Facial Action Unit Fusion," propose an architecture that simultaneously learns a facial graph representation and an AUs matrix, then merges the two channels to classify micro-expressions. The architecture extracts features from MagNet \cite{oh2018} using geometric points for eyebrows and mouth, creating a 30-node facial graph. Features within these nodes are integrated through convolution, preserving spatial information. Concurrently, nine AUs from eyebrow and mouth areas are integrated into a GCN to produce an AU feature matrix. Finally, a fusion mechanism combines the facial graph representation and the AU matrix for the final emotion classification. 

"Learning Dynamic Relationships for Facial Expression Recognition Based on Graph Convolutional Network" by X. Jin \textit{et al.} \cite{jin2021learning}, proposes a Dual Dynamic Relational GCN (DDRGCN). This architecture involves creating a region of interest graph directly linked to certain AUs (i.e., facial areas). These graphs are then used as input to a GCN to classify emotions. The resulting network surpasses existing lightweight networks in terms of accuracy, model size, and speed by focusing on the dynamic relationships inherent in facial expressions. 

J. Zhou \textit{et al.} \cite{zhou2020learning}, in "Learning the Connectivity: Situational Graph Convolution Network for Facial Expression Recognition," in 2020, introduced a framework using graphs for emotion recognition. They introduce a Situational Link Generation Module (SLGM), enabling a dual-stream architecture. These two streams are used in a Situational Graph Convolution Network (SGCN) to classify the emotions in a video. This architecture is noteworthy because it introduces both a customized graph creation and a custom GCN. The results presented were superior or close to the state-of-the-art at the time of publication. Ultimately, the architecture demonstrated robustness to varied experimental conditions, such as face occlusion. 

A. Panagiotis \textit{et al.} \cite{panagiotis2021exploiting}, with "Exploiting Emotional Dependencies with Graph Convolutional Networks for Facial Expression Recognition" introduce a multi-task learning architecture exploiting a classification model and a regression model. The former serves for basic emotion classification, and the latter for valence/arousal regression. While the feature extraction from images is performed with a CNN, the creation of classification and regression models is done using a GCN. They learn the relationships between different emotions and valence/arousal values using a GCN to obtain classification and regression models.

The work of N. Xie \textit{et al.} \cite{xie2023}: "Attention-Based Global-Local GRL for Dynamic Facial Expression Recognition" introduces a dual-stream method, extracting high-level visual features on one side using a Global Spatial-Temporal Graph (GSTG) and an attention mechanism. On the other side, it extracts local and geometric data via a Local Spatial-Temporal Graph (LSTG) from facial key points. The streams are then merged for emotion classification. This pipeline demonstrates robust performance even under challenging conditions such as occlusions and low lighting. 

With "Spatial-Temporal Graphs Plus Transformers for Geometry-Guided Facial Expression Recognition," R. Zhao \textit{et al.} \cite{zhao2022spatial}, present an architecture for geometry-guided FER. Similarly to the dual-stream method, this approach consists of several important steps and initially involves creating two distinct graphs per video. The first aims to extract visual information by obtaining super-regions around facial landmarks, and the second aims to extract coordinates from these landmarks. The information from these graphs is then merged and passed through spatial and temporal GCNs. The extracted features are finally passed through a transformer module to introduce an attention concept. In addition to its state-of-the-art results, an artificial occlusion test was conducted, demonstrating the robustness of this architecture. 

These various approaches highlight a trend towards more generalist and nuanced models in GRL for FER. By delving deeper into the spatio-temporal aspects of facial expressions, and using the capabilities of graph representation extensively, researchers are paving the way towards systems that are more accurate and capable of understanding the subtleties of human emotions. These advancements signal progress in applying GRL in FER and underscore the potential for these technologies to evolve further.

\section{Research Opportunities}
Many areas in the FER field can be explored in greater depth. These include optimizing recognition under various lighting conditions, managing occlusions, and adapting to variations in facial expressions and head poses. Particular interest should be given to exploring the capabilities of GRL to enhance the modeling of complex and dynamic relationships between facial features, thereby improving the models' capabilities to interpret expressions accurately in diverse contexts. Additional work can focus on developing innovative approaches to handle spatio-temporal data more efficiently, enhancing the encoding and interpretation of facial expressions across different head poses, adopting advanced semi-supervised and unsupervised learning methods, and optimizing data augmentation techniques for graphs. These initiatives offer promising avenues to enhance the generalization and robustness of FER models. 

To summarize, we believe that research should focus on leveraging GRL to model complex and dynamic graphs for more robustness, develop efficient methods for handling data, and adopt advanced learning techniques to enhance the robustness of FER models.
\section{Conclusion}
In this article, we introduced the notions of FER, graph theory, graph representation, and GRL. We then presented different databases related to FER, providing an overview of the different sets available to train, validate, and test GRL models. These databases range from lab collections to more realistic sets captured in natural environments, primarily featuring macro-expressions. We also synthesized recent advances and approaches in the field, highlighting the performance of various GRL methods on these datasets. The use of innovative approaches to represent data, such as spatio-temporal graphs has led to significant improvements in accuracy, robustness, and generalization of FER systems. These advancements show the rapid evolution of the field, the capabilities of graphs to create complex and clear relations between nodes offer good perspectives for the research on FER. 
\section{Acknowledgements}
We would like to express our gratitude to IRCICA for its indispensable support and considerable contribution which were fundamental to the completion of this article.
\newpage
\onecolumn
\twocolumn
\bibliographystyle{style/IEEEtran}
\bibliography{style/IEEEabrv,bibli}
\end{document}